\documentclass[11pt,a4paper]{article}
\usepackage[hyperref]{emnlp2020}
\usepackage{times}
\usepackage{latexsym}

\usepackage{microtype}

\aclfinalcopy 

\usepackage{amsmath}
\usepackage{multirow}
\usepackage{amssymb}
\usepackage{mathtools}
\usepackage{booktabs}
\usepackage{graphicx}
\usepackage{subcaption}

\DeclarePairedDelimiter{\nint}\lfloor\rceil

\title{Fully Quantized Transformer for Machine Translation}

\author{
  Gabriele Prato \\
  Mila, Universit\'{e} de Montr\'{e}al\\
  \texttt{pratogab@mila.quebec} \\
  \And
  Ella Charlaix \\
  Huawei Noah's Ark Lab \\
  \texttt{ella.charlaix@huawei.com}
  \AND
  Mehdi Rezagholizadeh \\
  Huawei Noah's Ark Lab \\
  \texttt{mehdi.rezagholizadeh@huawei.com}
}

\date{}

\begin{document}
\maketitle
\begin{abstract}
State-of-the-art neural machine translation methods employ massive amounts of parameters. Drastically reducing computational costs of such methods without affecting performance has been up to this point unsuccessful. To this end, we propose FullyQT: an all-inclusive quantization strategy for the Transformer. To the best of our knowledge, we are the first to show that it is possible to avoid any loss in translation quality with a fully quantized Transformer. Indeed, compared to full-precision, our 8-bit models score greater or equal BLEU on most tasks. Comparing ourselves to all previously proposed methods, we achieve state-of-the-art quantization results.
\end{abstract}

\section{Introduction}
The idea of using neural networks for machine translation was only recently proposed \citep{kalchbrenner-blunsom-2013-recurrent-continuous,NIPS2014_5346,cho-etal-2014-properties}. Nonetheless, the approach became the state-of-the-art in the field \citep{2017arXiv171102132A,2018arXiv180600187O,2018arXiv180809381E}. A key element of this success was to allow the decoder to attend to all hidden states of the encoder \citep{2014arXiv1409.0473B}. A few variations to this additive attention mechanism have been proposed, such as multiplicative and self-attention \citep{2015arXiv150804025L,2016arXiv160106733C,2017arXiv170303130L}. The latter formed the basis of the Transformer network \cite{transformer}, which achieved state-of-the-art results in machine translation. Inspiring a new wave of work, numerous natural language processing tasks reached new heights \citep{BERT,MTDNN}. Unfortunately, these models use an enormous amount of parameters. Inference on resource-limited hardware such as edge-devices is thus impractical.

A solution to reduce the computational burden of these networks is to lower numerical precision. Consequently, numerical values can be represented using fewer bits \citep{229903,182695}. This method called quantization has the advantage of providing good compression rates with minimal loss in accuracy. It is also conveniently supported by most hardware. Properly quantizing the Transformer would allow computational speed gains at inference, as well as deployment on more constrained devices.

In this work, we propose a quantization-aware training strategy for the entire Transformer architecture. Our method is easy to implement and results are consistent with the full-precision Transformer. We test our approach on multiple translation tasks such as WMT14 EN-FR and WMT14 EN-DE and obtain state-of-the-art quantization results. In comparison with full-precision, our quantized models score equal or higher BLEU on most tasks. We are, to the best of our knowledge, the first to show that the Transformer architecture can be fully quantized without impairing translation quality. We also perform an ablation study and show that quantizing specific components of the Transformer improves BLEU score.

\section{Background}
In this section, we review a broad spectrum of quantization and pruning methods for neural network compression.

\subsection{Quantization}
Over the years, a large range of methods have been proposed to quantize neural networks. These include, among many others, binary \citep{2016arXiv160202830C}, ternary \citep{2015arXiv151003009L,2016arXiv160504711L}, uniform \citep{2017arXiv171205877J} and learned \citep{2018arXiv180710029Z} quantization. These methods can be universally applied to any type of neural network. To maintain performance though, specific architectures usually require custom tailored quantization schemes.

Several recent work explore recurrent neural network \citep{rnns} quantization. \citet{2016arXiv160806902O} propose an exponential quantization method for RNN weights. They find ternary and exponential quantization to work well on language modeling and speech recognition, while binary weights seemed ineffective. \citet{2016arXiv160907061H} quantize weights and activations of both RNNs and LSTMs \citep{lstms} to 2, 4 and 6-bit. Meanwhile, \citet{2016arXiv161110176H} propose modifications to the gates and interlinks of quantized LSTM and GRU \citep{2014arXiv1406.1078C} cells, as well as a balanced quantization method for weights. \citet{2016arXiv160908144W} successfully quantize a stacked sequence-to-sequence LSTM to 8-bit without any loss in translation quality. Most recently, \citet{NIPS2018_7341} propose applying different quantization methods for different RNN components.

With regards to CNNs \citep{6795724}, various works have also explored quantizing these models. \citet{2014arXiv1412.6115G} compare matrix factorization, binarization, $k$-means clustering, product quantization and residual quantization of CNNs. \citet{2015arXiv151206473W} apply quantization to both kernels and fully connected layers of convolutional neural networks. \citet{2016arXiv160305279R} propose using binary weighted filters on AlexNet \citep{NIPS2012_4824}. Testing their method on ImageNet, they show classification accuracy to be on par with full-precision. For faster inference and training, \citet{2016arXiv160606160Z} use low bitwidth weights, activations and gradients on CNNs.

Quantization has been applied in tandem with other compression methods. \citet{2015arXiv151000149H} combine pruning, quantization, weight sharing and Huffman coding. In another line of work, \citet{2018arXiv180205668P} employ quantization with knowledge distillation \citep{2015arXiv150302531H} for higher compression rates. Moreover, \citet{NIPS2018_8295} blend quantization with block based low-rank matrix approximation of embeddings.

\subsection{Pruning}
The pruning of neural networks for model compression has also been largely explored. \citet{NIPS1989_250} were the first to propose a Hessian based method to prune neural net weights. \citet{NIPS1993_749} later improved the method. More recently, \citet{2016arXiv160609274S} show that pruning a fully trained model and then retraining it can increase performance over the original non-pruned model. Gradually pruning in tandem with training has also been shown to increase performance \citep{2017arXiv171001878Z}. To avoid sparse matrices, \citet{2017arXiv170806519L} prune nodes instead of weights. They apply a penalty in the loss on the $\gamma$ parameters of batch normalization layers. With a similar objective, \citet{2017arXiv171102782N} make better use of hardware by applying pruning and weight decay in blocks to minimize the number of loaded weight matrix chunks.

Similarly to quantization, pruning methods have also been adapted to specific architectures. \citet{Liu_2015_CVPR} propose an efficient sparse matrix multiplication algorithm for CNNs. As for RNNs, \citet{2017arXiv170405119N} show sparse pruning to work well on the architecture. In order to maintain dimension consistency, \citet{2017arXiv170905027W} propose to prune all basic LSTM structures concurrently. Lastly, \citet{NIPS2018_8261} introduce simple recurrent units (SRUs) for easy pruning of RNNs.

\section{FullyQT}
\subsection{Quantization Methodology}
Our quantization scheme was chosen to be uniform, meaning that the step size between two quantized values is constant. This choice, which is an additional constraint, was made for practical reasons. It indeed simplifies all computations required during inference, enabling the exploitation of hardware resources more efficiently. If the performance with uniform quantization is already on par with full-precision, then more weighty methods are unnecessary. A brief overview of uniform quantization is given in this section. For more details, we refer the reader to \citet{2017arXiv171205877J}.

Given an element $x$ of a tensor $\mathbf X$, we apply the quantization function $\mathcal{Q}$:
{\fontsize{08}{11}\selectfont
\begin{gather}
    \mathcal{Q}(x) = \nint*{\frac{\mathrm{clamp}(x; x_{min}, x_{max}) - x_{min}}{s}} * s + x_{min} \\
    s = \frac{x_{max} - x_{min}}{2^k -1}
\end{gather}
}
where $x_{min}$ and $x_{max}$ defines the endpoints of the quantization interval. When quantization is applied to weights, these values are respectively $\min(\mathbf X)$ and $\max(\mathbf X)$. However, when quantization is applied to activations, those values are running estimates.
The latter are computed during training, where for every forward pass, the $x_{min}$ and $x_{max}$ variables are updated via an exponential moving average with a momentum of 0.9. The $\mathrm{clamp}$ function associates all values outside of the $[x_{min}, x_{max}]$ range to the closest endpoint and $\nint*{\cdot}$ represents rounding to the nearest integer. The value $k$ is simply the bit precision. For example, in the context of 8-bit quantization, $k = 8$.

During backpropagation, we use the straight-through estimator \citep{ste} and set the gradients of clamped values to zero. Once training is finished, $s$ and $x_{min}$ are frozen along with the weights.

\subsection{What to Quantize}
We choose to quantize all operations which can provide a computational speed gain at inference. In this regard, we quantize all matrix multiplications, meaning that the inputs and weights of MatMuls will both be $k$-bit quantized. The other operations we quantize are divisions, but only if both the numerator and denominator are second or higher rank tensors. For all other operations, such as sums, the computational cost added by the quantization operation outweighs the benefit of performing the operation with reduced precision. Hence, we do not quantize such operations.

More precisely, we quantize all weights of the Transformer, excluding biases. The latter are summed with the INT32 output of matrix multiplications and thus provide no additional computational efficiency from being quantized. Furthermore, the memory space of biases is insignificant in comparison to the weight matrices, representing less than 0.1\% of total weights. For positional embeddings, these are fixed and can thus be quantized once prior to training. The $\gamma$ weights of LayerNorms are also quantized. As for activations, we quantize the sum of the input embeddings with the positional encodings in both the encoder and decoder. In the Multi-Head Attention, we quantize the $(Q, K, V)$ input, the softmax's numerator, the softmax's denominator, the softmax's output and the Scaled Dot-Product Attention's output. At inference, the softmax does not need to be computed in full-precision. Indeed, the exponential function can instead be replaced with a step function. For the position-wise feed-forward networks, we quantize the output of the ReLUs and of the feed-forwards themselves. Finally, for all LayerNorms, we quantize the numerator $x - \mu$, the denominator $\sqrt{\sigma^2 + \epsilon}$, their quotient and the output of the LayerNorm. A visual guide is provided in appendix \ref{sec:visual_guide}.

\subsection{Bucketing}
Instead of using a single set of $(s, x_{min})$ per quantized tensor, we can quantize subsets of the latter with each its own set of $(s, x_{min})$ \citep{2016arXiv161002132A}. Even though this adds more scalars, the memory cost is insignificant overall. Furthermore, the added flexibility can greatly alleviate the precision loss resulting from all values being mapped to a single low numerical precision domain.

We use this bucketing method for all weight matrices, with a number of subset equal to the output dimension. For activations, we use bucketing when quantizing: the sum of input embeddings with the positional encoding, the $Q, K, V$ inputs, the Scaled Dot-Product Attention's output, the feed-forward's output, the LayerNorm's numerator, quotient and output.

\subsection{Dealing with Zeros}
Unlike \citet{2017arXiv171205877J}, we do not nudge the domain so that the zero value gets perfectly mapped. The only zero values which we have to deal with are the padding, the Softmax numerator and output, the output of ReLU layers and dropouts. Since padding has no effect on the final output, we completely ignore these values when quantizing. For ReLUs and the Softmax's numerator and output, we fix their $x_{min}$ to 0, which guarantees the perfect mapping of the value. Finally, quantization is applied before any dropout operation. Indeed, even though the zeros added to the output of the quantization layer might not be part of the domain, this only happens during training.


\section{Related Work} \label{sec:related_work}
Recently, simple quantization solutions have been applied to the Transformer. \citet{15763707} apply $k$-means quantization and binarization with two centroids over the weights of the network. For both methods, a look up table associated with each quantized layer is used to map indices to their corresponding centroids. Similarly, \citet{15742249} compares binary, 4 and 8-bit uniform quantization of the Transformer weights. A big disadvantage with quantizing only the weights of a network is that operations must still be performed in full-precision. Even though the parameters' memory usage is reduced, these constantly have to be converted back to full-precision. Achieving quantization of both weights and activations is much more beneficial. The first attempt at doing so for the Transformer applies 8-bit quantization on weights and inputs of feed forward layers and binarizes the $(Q, K)$ input of the Multi-Head Attention \citep{15848474}. The scaling factor $\sqrt{d_k}$ is approximated by a constant which can be computed as a right bitshift. The method resulted in a huge drop in translation accuracy. Achieving better performance, \citet{2019arXiv190600532B} quantize certain MatMul operations and use the KL divergence to estimate the most suited parameters for each quantization range. They restrain from quantizing all MatMuls, reporting poorer results in accuracy. Aside from translation, the concurrent work by \citet{2019arXiv191006188Z} quantizes the embedding and fully connected layers of BERT \cite{BERT}. The Softmax and LayerNorm operations are kept in full-precision. On the GLUE benchmark, their loss in accuracy is minimal compared to the original model.

All of these methods omit quantizing the whole Transformer architecture, resulting in suboptimal computational efficiency. Furthermore, these solutions all fail to avoid impairing translation quality. Our method achieves both.

\begin{table*}[!ht]
  \centering
  \resizebox{\textwidth}{!}{
  \begin{tabular}{lcccccc}
  \toprule
  \multirow{2}{*}{Method} & Fully & Size (Gb) & \multirow{2}{*}{Compr.} & \multicolumn{3}{c}{BLEU} \\
  & Quantized & [EN-DE, EN-FR] & & EN-DE (2014) & EN-FR & EN-DE (2017) \\
  \midrule
  \citet{transformer} & & [2.02, 1.94] & 1x & 27.3 & 38.1 & - \\
  \citet{15763707} & & 0.69 & 2.92x & - & - & 27.38 \\
  \citet{2019arXiv190600532B} & & $\geq$ 0.96 & $\leq$ 2.1x & 27.33 & - & - \\
  \citet{15742249} & & $\geq$ 0.51 & $\leq$ 3.99x & 26.94 & - & - \\
  FullyQT & \checkmark & [0.52, 0.50] & 3.91x & \textbf{27.60} & \textbf{39.91} & \textbf{27.60}\\
  \bottomrule
  \end{tabular}
  }
  \caption{\label{best_results}
  Our quantization strategy achieves better BLEU scores than all other quantization methods for the Transformer on the WMT14 EN-DE, WMT14 EN-FR and WMT17 EN-DE test set.}
\end{table*}

\begin{table*}[!ht]
  \resizebox{\textwidth}{!}{
  \begin{tabular}{llc|cccc|cccc}
  \toprule
  \multirow{2}{*}{Model} & \multirow{2}{*}{Method} & \multirow{2}{*}{Precision} & \multicolumn{4}{c}{EN-DE} & \multicolumn{4}{c}{EN-FR} \\
  & & & PPL & BLEU & Size (Gb) & Compr. & PPL & BLEU & Size (Gb) & Compr. \\
  \midrule
  Base & Baseline & 32-bit & 4.95 & 26.46 & 2.02 & 1x & 3.21 & 38.34 & 1.94 & 1x \\
  & Default Approach & 8-bit & 74.04 & 0.21 & 0.52 & 3.91x & \textit{nan} & 0 & 0.50 & 3.91x \\
  & Post-Quantization & 8-bit & 4.97 & 26.44 & 0.52 & 3.91x & 3.26 & 38.30 & 0.50 & 3.91x \\
  & FullyQT & 8-bit & 4.94 & 26.38 & 0.52 & 3.91x & 3.23 & \textbf{38.41} & 0.50 & 3.91x \\
  & Post-Quantization & 6-bit & 6.00 & 24.84 & 0.39 & 5.18x & 3.98 & 35.02 & 0.37 & 5.17x \\
  & FullyQT & 6-bit & 5.09 & \textbf{26.98} & 0.39 & 5.18x & 3.38 & 37.07 & 0.37 & 5.17x \\
  & FullyQT & 4-bit & 11.96 & 18.32 & 0.26 & 7.66x & 48.21 & 1.59 & 0.25 & 7.64x \\
  \midrule
  Big & Baseline & 32-bit & 4.38 & \textbf{27.13} & 6.85 & 1x & 2.77 & \textbf{40.54} & 6.69 & 1x \\
  & Post-Quantization & 8-bit & 4.27 & 26.55 & 1.74 & 3.95x & 2.78 & 39.78 & 1.69 & 3.95x \\
  & FullyQT & 8-bit & 4.57 & 26.96 & 1.74 & 3.95x & 2.80 & 40.25 & 1.69 & 3.95x \\
  & Post-Quantization & 6-bit & 5.12 & 24.86 & 1.31 & 5.24x & 3.08 & 37.92 & 1.28 & 5.24x \\
  & FullyQT & 6-bit & 4.78 & 26.76 & 1.31 & 5.24x & 2.87 & 39.59 & 1.28 & 5.24x \\
  & FullyQT & 4-bit & 33.11 & 10.22 & 0.88 & 7.79x & 42.42 & 2.81 & 0.86 & 7.79x \\
  \bottomrule
  \end{tabular}
  }
  \caption{\label{custom_quantization_vs_naive}
  Performance of our quantization method on the WMT14 EN-DE and WMT14 EN-FR test set for a fixed number of training steps.}
\end{table*}

\section{Experiments}
In this section, we present the results of our full quantization scheme on various tasks. We first compare our method on a machine translation setup. Then we present the results of numerous ablation studies. We also compare the impact of delaying quantization on translation quality. Finally, we evaluate our method on two language model tasks.

\subsection{Full Quantization} \label{sec:full_quantization}
We apply our quantization strategy on both the base and big Transformer \citep{transformer}. The training setup of all presented models is the same as in the original paper, with the exception that the dropout ratio is set to 0.1 in all cases. We refer readers to the original paper for experimental details. Our models were first evaluated on the WMT 2014 / 2017 English-to-German and WMT 2014 English-to-French translation tasks. Section \ref{sec:additional_datasets} contains results on additional language datasets. Reported perplexity is per token and BLEU was measured with \texttt{multi-bleu.pl}\footnote{\label{note2}\url{https://github.com/moses-smt/mosesdecoder/blob/master/scripts/generic/multi-bleu.perl}} on the \texttt{newstest2014}\footnote{\label{note3}\url{https://www.statmt.org/wmt14/translation-task.html}} test set. We used beam search with a beam size of 4 and a length penalty of 0.6. Unlike \citet{transformer}, no checkpoint averaging was performed.

We compare our results with the original Transformer and other 8-bit quantization methods in Table \ref{best_results}. All models are base Transformers. Original uncompressed size is the same in all cases. Most work do not report their compressed model size. For those, we give lower bounds based on their reports. Our BLEU score was computed on the test set using the checkpoint with the highest validation accuracy over 2 million training steps. Validation was computed every training epoch. Models were trained once. Our objective was to train quantized models up to convergence. Very similar BLEU scores can be obtained with much fewer training (see below). As for other methods, \citet{15763707} retrain for 10k steps a 200k steps pretrained Transformer. \citet{15742249} also does the same but does not mention the number of retraining steps. \citet{2019arXiv190600532B} and the original Transformer paper both do not mention the number of training steps. Out of all methods, we are the only one quantizing every component of the model (see section \ref{sec:related_work} for details).

In Table \ref{custom_quantization_vs_naive}, we show performance of our method on the WMT14 EN-DE and WMT14 EN-FR for a fixed amount of training steps. We compare our results with two full-precision Transformers: base and big variants. We also evaluate two other quantization approaches. The first one is the "default" approach, which is to naively quantize every possible operation. The second approach applies our quantization strategy post-training (see section \ref{delaying_quantization}). In all cases except for post-quantization, BLEU was computed on the test set using the checkpoint which scored the highest accuracy on the validation set. Towards the end of training, we ran one validation epoch for every 100 training steps. Baselines and FullyQT 8-bit results were averaged over 5 trials. Standard deviation of the BLEU scores did not seem higher for any method and ranged between 0.09 and 0.51. Training with quantization was about twice as slow as with the baselines. As for post-training quantization, the BLEU score was computed on the test set using the best validation performance out of 20 trials. The default approach's \textit{nan} in the EN-FR task is due to numerical instability. By quantizing every operation, zeros in the LayerNorm's denominator are more frequent.

Looking at all conducted experiments, including section \ref{delaying_quantization} and appendix \ref{sec:additional_datasets}, translation quality of the 8-bit FullyQT models seems to be on par with full-precision. Most of the time, the highest BLEU was scored by the quantized model. For example in the case of WMT14 EN-DE, the maximum BLEU FullyQT base 8-bit obtained was 26.98, while the baseline's highest was 26.64. As for the big models, the max FullyQT scored was 27.95, whereas the baseline's was 27.43. We looked at training and validation curves to see if quantization had any effect, but saw no discernible difference.

All models use full-precision biases, $s$ and $x_{min}$. This amounts to 11.61 Mb in the base models and 23.15 Mb in the big models. In the case of 8-bit, these represent less than 2.35\% of the total size. Without bucketing, this would amount to 2.18 Mb and 4.35 Mb respectively. We believe the small increase in model size to be worth it. Indeed, in section \ref{sec:ablation_study}, we show that training without bucketing leads to poorer translation.

Although 6-bit quantization seems to perform well, the compression advantage over 8-bit is usually lost. Most hardware store INT6 using either 8 or 32 bits. Dedicated hardware is needed to get the full compression advantage. Unless 6-bit quantization results in better models, 8-bit seems like the best choice for most hardware.

\begin{table*}[!ht]
\centering
\resizebox{0.75\textwidth}{!}{
\begin{tabular}{clcccc}
\toprule
\multirow{2}{*}{Module} & \multirow{2}{*}{Quantized Activation} & \multicolumn{2}{c}{No Bucketing} & \multicolumn{2}{c}{Bucketing} \\
& & PPL & BLEU & PPL & BLEU \\
\midrule
Encoder & (Input Embedding + Positional Encoding) & 3.20 & 38.61 & 3.20 & 39.08 \\
\midrule
Decoder & (Input Embedding + Positional Encoding) & 3.20 & 39.35 & 3.20 & 39.36 \\
\midrule
\multirow{2}{*}{\shortstack[c]{Multi-Head\\ Attention}} & Input $(Q,K,V)$ & 3.21 & 39.06 & 3.21 & 39.29 \\
& LayerNorm Output & 3.21 & 39.09 & 3.20 & 38.78 \\
\midrule
\multirow{4}{*}{\shortstack[c]{Scaled\\ Dot-Product\\ Attention}} & Softmax Numerator & 3.20 & 39.32 & 3.21 & 39.01 \\
& Softmax Denominator & 3.21 & 39.35 & 3.21 & 39.11 \\
& Softmax Output & 3.22 & 39.41 & 3.22 & 38.87 \\
& Output & 3.21 & 38.73 & 3.21 & 39.02 \\
\midrule
\multirow{3}{*}{Feed-forward} & ReLU Output & 3.21 & 39.43 & 3.22 & 38.93 \\
& Feed-forward Output & 3.54 & 38.03 & 3.20 & 39.27 \\
& LayerNorm Output & 3.21 & 38.67 & 3.21 & 39.04 \\
\midrule
\multirow{3}{*}{LayerNorm} & Numerator & 3.53 & 37.75 & 3.21 & 38.86 \\
& Denominator & 1748 & 0 & - & - \\
& Quotient & 3.22 & 38.97 & 3.21 & 39.02 \\
\bottomrule
\end{tabular}
}
\caption{\label{ablation_study}
Effect of quantizing single activations of the Transformer. Results are on the WMT14 EN-FR test set.}
\end{table*}

\begin{table*}[!ht]
  \centering
  \resizebox{0.6\textwidth}{!}{
  \begin{tabular}{lcc}
  \toprule
  Method & PPL & BLEU \\
  \midrule
  No Bucketing & 3.49 & 37.14 \\
  No Gradient Clipping & 2549.30 & 0 \\
  No LayerNorm Denominator Quantization & 3.22 & 38.29 \\
  8-bit Quantized Weights, Full-precision Activations & 3.20 & 38.36 \\
  \bottomrule
  \end{tabular}
  }
  \caption{\label{tab:more_ablation}
  Variations to our quantization scheme evaluated on the WMT14 EN-FR translation task.}
\end{table*}

\subsection{Ablation Studies} \label{sec:ablation_study}
To better understand which operations are more sensitive to quantization, we evaluate such effect on single operations of the Transformer. By this, we mean quantizing the operation of a module for all Transformer layers. Table \ref{ablation_study} shows results on the WMT14 EN-FR translation task for 8-bit precision. The effect of bucketing was also evaluated. BLEU was computed on the test set after 100k steps of training. In 24 out of 27 experiments, performance was better than our full-precision baseline of 38.34 BLEU. Solely quantizing the LayerNorm's denominator with no bucketing results in poor performance. The latter also cannot be bucketed since all dimensions of the variance tensor vary per batch. To successfully quantize this element without causing any loss in performance, we suspect quantizing other elements in the network helps.

To further validate our quantization scheme, we evaluated four models trained with alterations to our design choices. Results on the WMT14 EN-FR task are presented in Table \ref{tab:more_ablation}. All models are 8-bit quantized base Transformers. Training procedure is the same as in section \ref{sec:full_quantization}.

\begin{table}[!ht]
  \centering
  \resizebox{\columnwidth}{!}{
  \begin{tabular}{lcccc}
  \toprule
  Quantization Start & \multicolumn{2}{c}{EN-DE} & \multicolumn{2}{c}{EN-FR} \\
  (training step) & PPL & BLEU & PPL & BLEU \\
  \midrule
  Never quantized & 4.95 & 26.46 & 3.21 & 38.34 \\
  100 & 4.67 & \textbf{26.98} & 3.23 & 38.55 \\
  10000 & 4.99 & 26.63 & 3.21 & \textbf{38.62} \\
  50000 & 4.98 & 26.84 & 3.21 & 38.50 \\
  80000 & 5.03 & 26.41 & 3.21 & 38.43 \\ 
  Post-Quantization & 4.45 & 25.50 & 3.22 & 37.96 \\
  \bottomrule
  \end{tabular}
  }
  \caption{\label{quantization_start}
  Impact of delaying quantization. Results are on the WMT14 EN-DE and WMT14 EN-FR test set.}
\end{table}

\begin{table*}[!ht]
  \centering
  \resizebox{0.6\textwidth}{!}{
  \begin{tabular}{lcccccc}
  \toprule
  \multirow{2}{*}{Precision} & \multirow{2}{*}{Size (Mb)} & \multirow{2}{*}{Compression} & \multicolumn{2}{c}{WikiText-2} & \multicolumn{2}{c}{WikiText-103} \\
  & & & Loss & PPL & Loss & PPL \\
  \midrule
  32-bit & 243.04 & 1x & 5.65 & 284.15 & 5.91 & \textbf{369.20} \\
  8-bit & 61.93 & 3.92x & 5.64 & 282.67 & 5.94 & 377.79 \\
  6-bit & 46.75 & 5.20x & 5.64 & \textbf{281.48} & 5.93 & 376.44 \\
  4-bit & 31.57 & 7.70x & 5.65 & 284.26 & 5.94 & 378.67 \\
  \bottomrule
  \end{tabular}
  }
  \caption{\label{tab:language_modeling}
  Evaluation of our quantization method on the WikiText-2 and WikiText-103 language modeling tasks.}
\end{table*}

\subsection{Delaying Quantization} \label{delaying_quantization}
Our method's goal is to increase computational efficiency when inferring with the Transformer. To this end, our quantization scheme only requires us to learn $s$ and $x_{min}$. Although we do so throughout the whole training, this is not a necessity. Quantization could also be applied later during training. Results for different starting points are compared in Table \ref{quantization_start}. The earliest we start quantizing is at 100 steps, since we need at least a few steps to assess the running estimates. All models were evaluated on the WMT14 EN-DE and WMT14 EN-FR translation tasks. BLEU was measured on the test set using the checkpoint which scored the highest accuracy on the validation set during training. Validation was computed every 100 training steps towards the end of training. From our observed results, quantizing the model early on seems preferable.

Learning quantization parameters adds a significant computational cost during training. A major advantage to delaying quantization is to perform more training steps in the same given amount of time. Therefore, when training time is a constraint, a possible strategy is to train a model without quantization, perform more training steps and finally post-quantize the model. By the latter, we mean keeping all weights fixed and compute the $s$ and $x_{min}$ over a few hundred steps. This is another advantage, since many trials can be performed in search of the best performing candidate. We found post-quantization BLEU scores to vary by about 0.2 BLEU.

\subsection{Language Modeling}
To evaluate if our quantization scheme generalizes well to other tasks, we evaluate it on two language modeling datasets: WikiText-2 and WikiText-103. As the setup, we use PyTorch's language modeling toy example\footnote{\label{note4}\url{https://github.com/pytorch/examples/tree/master/word_language_model}}. The task consists of predicting the sequence $\{x_{t+1}, \cdots, x_{t+n+1}\}$ from the input sequence $\{x_{t}, \cdots, x_{t+n}\}$. We trained four Transformer models, each with different precision. All models consist of two Transformer encoder layers, with the embedding and hidden size set to 200. Multi-Head Attention has two heads with key and value size 64. The final word projection layer's weights are shared with the embedding layer. Models were trained for 10 epochs with a batch size of 20 and sequence length of 35. Learning rate is set to 5, dropout to 0.2 and gradient clipping to 0.25. Loss is computed on every element of the output sequence. Results are presented in Table \ref{tab:language_modeling}. Validation was computed every epoch to determine the best candidate. Loss and perplexity are computed on the test set and averaged over 10 trials for WikiText-2 and 3 trials for WikiText-3. See footnote \ref{note4} for any extra details.

\section{Conclusion}
We proposed a full quantization strategy for the Transformer architecture. Our objective was to exploit hardware resources as efficiently as possible, quantizing all operations which could provide a computational speed gain.

With FullyQT, we achieve higher BLEU scores than all other quantization methods for the Transformer on multiple translation tasks and avoid any loss in BLEU compared to full-precision. Specifically, out of 35 experiments, 8-bit quantization performed better than full-precision in 21 cases.

If instead of minimizing inference time, one wants to maximize translation accuracy, then applying quantization to only certain components of the Transformer seems to be the best option. Indeed, our ablation study showed than BLEU score could increase even more when only specific elements of the Transformer were quantized. Further gains might be possible, but supplementary experiments would be necessary to determine the best combination.

We are very excited about the possibilities this work opens and plan on applying our method to other tasks. We also intend to extend our work to variations of the Transformer, as well as further exploring the compression of these networks.

\bibliography{references}

\begin{thebibliography}{55}
\expandafter\ifx\csname natexlab\endcsname\relax\def\natexlab#1{#1}\fi

\bibitem[{{Ahmed} et~al.(2017){Ahmed}, {Shirish Keskar}, and
  {Socher}}]{2017arXiv171102132A}
Karim {Ahmed}, Nitish {Shirish Keskar}, and Richard {Socher}. 2017.
\newblock \href {http://arxiv.org/abs/1711.02132} {{Weighted Transformer
  Network for Machine Translation}}.
\newblock \emph{arXiv e-prints}, page arXiv:1711.02132.

\bibitem[{{Alistarh} et~al.(2016){Alistarh}, {Grubic}, {Li}, {Tomioka}, and
  {Vojnovic}}]{2016arXiv161002132A}
Dan {Alistarh}, Demjan {Grubic}, Jerry {Li}, Ryota {Tomioka}, and Milan
  {Vojnovic}. 2016.
\newblock \href {http://arxiv.org/abs/1610.02132} {{QSGD:
  Communication-Efficient SGD via Gradient Quantization and Encoding}}.
\newblock \emph{arXiv e-prints}, page arXiv:1610.02132.

\bibitem[{{Bahdanau} et~al.(2014){Bahdanau}, {Cho}, and
  {Bengio}}]{2014arXiv1409.0473B}
Dzmitry {Bahdanau}, Kyunghyun {Cho}, and Yoshua {Bengio}. 2014.
\newblock \href {http://arxiv.org/abs/1409.0473} {{Neural Machine Translation
  by Jointly Learning to Align and Translate}}.
\newblock \emph{arXiv e-prints}, page arXiv:1409.0473.

\bibitem[{{Bhandare} et~al.(2019){Bhandare}, {Sripathi}, {Karkada}, {Menon},
  {Choi}, {Datta}, and {Saletore}}]{2019arXiv190600532B}
Aishwarya {Bhandare}, Vamsi {Sripathi}, Deepthi {Karkada}, Vivek {Menon}, Sun
  {Choi}, Kushal {Datta}, and Vikram {Saletore}. 2019.
\newblock \href {http://arxiv.org/abs/1906.00532} {{Efficient 8-Bit
  Quantization of Transformer Neural Machine Language Translation Model}}.
\newblock \emph{arXiv e-prints}, page arXiv:1906.00532.

\bibitem[{Chen et~al.(2018)Chen, Si, Li, Chelba, and Hsieh}]{NIPS2018_8295}
Patrick Chen, Si~Si, Yang Li, Ciprian Chelba, and Cho-Jui Hsieh. 2018.
\newblock Groupreduce: Block-wise low-rank approximation for neural language
  model shrinking.
\newblock In S.~Bengio, H.~Wallach, H.~Larochelle, K.~Grauman, N.~Cesa-Bianchi,
  and R.~Garnett, editors, \emph{Advances in Neural Information Processing
  Systems 31}, pages 10988--10998. Curran Associates, Inc.

\bibitem[{{Cheng} et~al.(2016){Cheng}, {Dong}, and
  {Lapata}}]{2016arXiv160106733C}
Jianpeng {Cheng}, Li~{Dong}, and Mirella {Lapata}. 2016.
\newblock \href {http://arxiv.org/abs/1601.06733} {{Long Short-Term
  Memory-Networks for Machine Reading}}.
\newblock \emph{arXiv e-prints}, page arXiv:1601.06733.

\bibitem[{{Cheong} and {Daniel}(2019)}]{15763707}
Robin {Cheong} and Robel {Daniel}. 2019.
\newblock \href
  {https://web.stanford.edu/class/cs224n/reports/custom/15763707.pdf}
  {{transformers.zip: Compressing Transformers with Pruning and Quantization}}.
\newblock Technical report, Stanford University, Stanford, California.

\bibitem[{Cho et~al.(2014)Cho, van Merrienboer, Bahdanau, and
  Bengio}]{cho-etal-2014-properties}
Kyunghyun Cho, Bart van Merrienboer, Dzmitry Bahdanau, and Yoshua Bengio. 2014.
\newblock \href {https://doi.org/10.3115/v1/W14-4012} {On the properties of
  neural machine translation: Encoder{--}decoder approaches}.
\newblock In \emph{Proceedings of {SSST}-8, Eighth Workshop on Syntax,
  Semantics and Structure in Statistical Translation}, pages 103--111, Doha,
  Qatar. Association for Computational Linguistics.

\bibitem[{{Cho} et~al.(2014){Cho}, {van Merrienboer}, {Gulcehre}, {Bahdanau},
  {Bougares}, {Schwenk}, and {Bengio}}]{2014arXiv1406.1078C}
Kyunghyun {Cho}, Bart {van Merrienboer}, Caglar {Gulcehre}, Dzmitry {Bahdanau},
  Fethi {Bougares}, Holger {Schwenk}, and Yoshua {Bengio}. 2014.
\newblock \href {http://arxiv.org/abs/1406.1078} {{Learning Phrase
  Representations using RNN Encoder-Decoder for Statistical Machine
  Translation}}.
\newblock \emph{arXiv e-prints}, page arXiv:1406.1078.

\bibitem[{{Courbariaux} et~al.(2016){Courbariaux}, {Hubara}, {Soudry},
  {El-Yaniv}, and {Bengio}}]{2016arXiv160202830C}
Matthieu {Courbariaux}, Itay {Hubara}, Daniel {Soudry}, Ran {El-Yaniv}, and
  Yoshua {Bengio}. 2016.
\newblock \href {http://arxiv.org/abs/1602.02830} {{Binarized Neural Networks:
  Training Deep Neural Networks with Weights and Activations Constrained to +1
  or -1}}.
\newblock \emph{arXiv e-prints}, page arXiv:1602.02830.

\bibitem[{{Devlin} et~al.(2018){Devlin}, {Chang}, {Lee}, and
  {Toutanova}}]{BERT}
Jacob {Devlin}, Ming-Wei {Chang}, Kenton {Lee}, and Kristina {Toutanova}. 2018.
\newblock \href {http://arxiv.org/abs/1810.04805} {{BERT: Pre-training of Deep
  Bidirectional Transformers for Language Understanding}}.
\newblock \emph{arXiv e-prints}, page arXiv:1810.04805.

\bibitem[{{Edunov} et~al.(2018){Edunov}, {Ott}, {Auli}, and
  {Grangier}}]{2018arXiv180809381E}
Sergey {Edunov}, Myle {Ott}, Michael {Auli}, and David {Grangier}. 2018.
\newblock \href {http://arxiv.org/abs/1808.09381} {{Understanding
  Back-Translation at Scale}}.
\newblock \emph{arXiv e-prints}, page arXiv:1808.09381.

\bibitem[{{Fan}(2019)}]{15742249}
Chaofei {Fan}. 2019.
\newblock \href
  {https://web.stanford.edu/class/cs224n/reports/custom/15742249.pdf}
  {{Quantized Transformer}}.
\newblock Technical report, Stanford University, Stanford, California.

\bibitem[{{Gong} et~al.(2014){Gong}, {Liu}, {Yang}, and
  {Bourdev}}]{2014arXiv1412.6115G}
Yunchao {Gong}, Liu {Liu}, Ming {Yang}, and Lubomir {Bourdev}. 2014.
\newblock \href {http://arxiv.org/abs/1412.6115} {{Compressing Deep
  Convolutional Networks using Vector Quantization}}.
\newblock \emph{arXiv e-prints}, page arXiv:1412.6115.

\bibitem[{{Han} et~al.(2015){Han}, {Mao}, and {Dally}}]{2015arXiv151000149H}
Song {Han}, Huizi {Mao}, and William~J. {Dally}. 2015.
\newblock \href {http://arxiv.org/abs/1510.00149} {{Deep Compression:
  Compressing Deep Neural Networks with Pruning, Trained Quantization and
  Huffman Coding}}.
\newblock \emph{arXiv e-prints}, page arXiv:1510.00149.

\bibitem[{Hassibi et~al.(1994)Hassibi, Stork, and Wolff}]{NIPS1993_749}
Babak Hassibi, David~G. Stork, and Gregory Wolff. 1994.
\newblock Optimal brain surgeon: Extensions and performance comparisons.
\newblock In J.~D. Cowan, G.~Tesauro, and J.~Alspector, editors, \emph{Advances
  in Neural Information Processing Systems 6}, pages 263--270. Morgan-Kaufmann.

\bibitem[{{He} et~al.(2016){He}, {Wen}, {Zhou}, {Wu}, {Yao}, {Zhou}, and
  {Zou}}]{2016arXiv161110176H}
Qinyao {He}, He~{Wen}, Shuchang {Zhou}, Yuxin {Wu}, Cong {Yao}, Xinyu {Zhou},
  and Yuheng {Zou}. 2016.
\newblock \href {http://arxiv.org/abs/1611.10176} {{Effective Quantization
  Methods for Recurrent Neural Networks}}.
\newblock \emph{arXiv e-prints}, page arXiv:1611.10176.

\bibitem[{{Hinton}(2012)}]{ste}
Geoffrey {Hinton}. 2012.
\newblock Neural networks for machine learning.
\newblock Coursera, video lectures.

\bibitem[{{Hinton} et~al.(2015){Hinton}, {Vinyals}, and
  {Dean}}]{2015arXiv150302531H}
Geoffrey {Hinton}, Oriol {Vinyals}, and Jeff {Dean}. 2015.
\newblock \href {http://arxiv.org/abs/1503.02531} {{Distilling the Knowledge in
  a Neural Network}}.
\newblock \emph{arXiv e-prints}, page arXiv:1503.02531.

\bibitem[{Hochreiter and Schmidhuber(1997)}]{lstms}
Sepp Hochreiter and Jürgen Schmidhuber. 1997.
\newblock \href {https://doi.org/10.1162/neco.1997.9.8.1735} {Long short-term
  memory}.
\newblock \emph{Neural computation}, 9:1735--80.

\bibitem[{{Hubara} et~al.(2016){Hubara}, {Courbariaux}, {Soudry}, {El-Yaniv},
  and {Bengio}}]{2016arXiv160907061H}
Itay {Hubara}, Matthieu {Courbariaux}, Daniel {Soudry}, Ran {El-Yaniv}, and
  Yoshua {Bengio}. 2016.
\newblock \href {http://arxiv.org/abs/1609.07061} {{Quantized Neural Networks:
  Training Neural Networks with Low Precision Weights and Activations}}.
\newblock \emph{arXiv e-prints}, page arXiv:1609.07061.

\bibitem[{{Jacob} et~al.(2017){Jacob}, {Kligys}, {Chen}, {Zhu}, {Tang},
  {Howard}, {Adam}, and {Kalenichenko}}]{2017arXiv171205877J}
Benoit {Jacob}, Skirmantas {Kligys}, Bo~{Chen}, Menglong {Zhu}, Matthew {Tang},
  Andrew {Howard}, Hartwig {Adam}, and Dmitry {Kalenichenko}. 2017.
\newblock \href {http://arxiv.org/abs/1712.05877} {{Quantization and Training
  of Neural Networks for Efficient Integer-Arithmetic-Only Inference}}.
\newblock \emph{arXiv e-prints}, page arXiv:1712.05877.

\bibitem[{Jordan(1990)}]{rnns}
Michael~I. Jordan. 1990.
\newblock Artificial neural networks.
\newblock chapter Attractor Dynamics and Parallelism in a Connectionist
  Sequential Machine, pages 112--127. IEEE Press, Piscataway, NJ, USA.

\bibitem[{Kalchbrenner and
  Blunsom(2013)}]{kalchbrenner-blunsom-2013-recurrent-continuous}
Nal Kalchbrenner and Phil Blunsom. 2013.
\newblock Recurrent continuous translation models.
\newblock In \emph{Proceedings of the 2013 Conference on Empirical Methods in
  Natural Language Processing}, pages 1700--1709, Seattle, Washington, USA.
  Association for Computational Linguistics.

\bibitem[{Krizhevsky et~al.(2012)Krizhevsky, Sutskever, and
  Hinton}]{NIPS2012_4824}
Alex Krizhevsky, Ilya Sutskever, and Geoffrey~E Hinton. 2012.
\newblock Imagenet classification with deep convolutional neural networks.
\newblock In F.~Pereira, C.~J.~C. Burges, L.~Bottou, and K.~Q. Weinberger,
  editors, \emph{Advances in Neural Information Processing Systems 25}, pages
  1097--1105. Curran Associates, Inc.

\bibitem[{{LeCun} et~al.(1989){LeCun}, {Boser}, {Denker}, {Henderson},
  {Howard}, {Hubbard}, and {Jackel}}]{6795724}
Y.~{LeCun}, B.~{Boser}, J.~S. {Denker}, D.~{Henderson}, R.~E. {Howard},
  W.~{Hubbard}, and L.~D. {Jackel}. 1989.
\newblock \href {https://doi.org/10.1162/neco.1989.1.4.541} {Backpropagation
  applied to handwritten zip code recognition}.
\newblock \emph{Neural Computation}, 1(4):541--551.

\bibitem[{LeCun et~al.(1990)LeCun, Denker, and Solla}]{NIPS1989_250}
Yann LeCun, John~S. Denker, and Sara~A. Solla. 1990.
\newblock Optimal brain damage.
\newblock In D.~S. Touretzky, editor, \emph{Advances in Neural Information
  Processing Systems 2}, pages 598--605. Morgan-Kaufmann.

\bibitem[{{Li} et~al.(2016){Li}, {Zhang}, and {Liu}}]{2016arXiv160504711L}
Fengfu {Li}, Bo~{Zhang}, and Bin {Liu}. 2016.
\newblock \href {http://arxiv.org/abs/1605.04711} {{Ternary Weight Networks}}.
\newblock \emph{arXiv e-prints}, page arXiv:1605.04711.

\bibitem[{{Lin} et~al.(2015){Lin}, {Courbariaux}, {Memisevic}, and
  {Bengio}}]{2015arXiv151003009L}
Zhouhan {Lin}, Matthieu {Courbariaux}, Roland {Memisevic}, and Yoshua {Bengio}.
  2015.
\newblock \href {http://arxiv.org/abs/1510.03009} {{Neural Networks with Few
  Multiplications}}.
\newblock \emph{arXiv e-prints}, page arXiv:1510.03009.

\bibitem[{{Lin} et~al.(2017){Lin}, {Feng}, {Nogueira dos Santos}, {Yu},
  {Xiang}, {Zhou}, and {Bengio}}]{2017arXiv170303130L}
Zhouhan {Lin}, Minwei {Feng}, Cicero {Nogueira dos Santos}, Mo~{Yu}, Bing
  {Xiang}, Bowen {Zhou}, and Yoshua {Bengio}. 2017.
\newblock \href {http://arxiv.org/abs/1703.03130} {{A Structured Self-attentive
  Sentence Embedding}}.
\newblock \emph{arXiv e-prints}, page arXiv:1703.03130.

\bibitem[{Liu et~al.(2015)Liu, Wang, Foroosh, Tappen, and
  Pensky}]{Liu_2015_CVPR}
Baoyuan Liu, Min Wang, Hassan Foroosh, Marshall Tappen, and Marianna Pensky.
  2015.
\newblock Sparse convolutional neural networks.
\newblock In \emph{The IEEE Conference on Computer Vision and Pattern
  Recognition (CVPR)}.

\bibitem[{{Liu} et~al.(2019){Liu}, {He}, {Chen}, and {Gao}}]{MTDNN}
Xiaodong {Liu}, Pengcheng {He}, Weizhu {Chen}, and Jianfeng {Gao}. 2019.
\newblock \href {http://arxiv.org/abs/1901.11504} {{Multi-Task Deep Neural
  Networks for Natural Language Understanding}}.
\newblock \emph{arXiv e-prints}, page arXiv:1901.11504.

\bibitem[{{Liu} et~al.(2017){Liu}, {Li}, {Shen}, {Huang}, {Yan}, and
  {Zhang}}]{2017arXiv170806519L}
Zhuang {Liu}, Jianguo {Li}, Zhiqiang {Shen}, Gao {Huang}, Shoumeng {Yan}, and
  Changshui {Zhang}. 2017.
\newblock \href {http://arxiv.org/abs/1708.06519} {{Learning Efficient
  Convolutional Networks through Network Slimming}}.
\newblock \emph{arXiv e-prints}, page arXiv:1708.06519.

\bibitem[{{Luong} et~al.(2015){Luong}, {Pham}, and
  {Manning}}]{2015arXiv150804025L}
Minh-Thang {Luong}, Hieu {Pham}, and Christopher~D. {Manning}. 2015.
\newblock \href {http://arxiv.org/abs/1508.04025} {{Effective Approaches to
  Attention-based Neural Machine Translation}}.
\newblock \emph{arXiv e-prints}, page arXiv:1508.04025.

\bibitem[{{Marchesi} et~al.(1993){Marchesi}, {Orlandi}, {Piazza}, and
  {Uncini}}]{182695}
M.~{Marchesi}, G.~{Orlandi}, F.~{Piazza}, and A.~{Uncini}. 1993.
\newblock \href {https://doi.org/10.1109/72.182695} {Fast neural networks
  without multipliers}.
\newblock \emph{IEEE Transactions on Neural Networks}, 4(1):53--62.

\bibitem[{{Narang} et~al.(2017{\natexlab{a}}){Narang}, {Elsen}, {Diamos}, and
  {Sengupta}}]{2017arXiv170405119N}
Sharan {Narang}, Erich {Elsen}, Gregory {Diamos}, and Shubho {Sengupta}.
  2017{\natexlab{a}}.
\newblock \href {http://arxiv.org/abs/1704.05119} {{Exploring Sparsity in
  Recurrent Neural Networks}}.
\newblock \emph{arXiv e-prints}, page arXiv:1704.05119.

\bibitem[{{Narang} et~al.(2017{\natexlab{b}}){Narang}, {Undersander}, and
  {Diamos}}]{2017arXiv171102782N}
Sharan {Narang}, Eric {Undersander}, and Gregory {Diamos}. 2017{\natexlab{b}}.
\newblock \href {http://arxiv.org/abs/1711.02782} {{Block-Sparse Recurrent
  Neural Networks}}.
\newblock \emph{arXiv e-prints}, page arXiv:1711.02782.

\bibitem[{{Ott} et~al.(2016){Ott}, {Lin}, {Zhang}, {Liu}, and
  {Bengio}}]{2016arXiv160806902O}
Joachim {Ott}, Zhouhan {Lin}, Ying {Zhang}, Shih-Chii {Liu}, and Yoshua
  {Bengio}. 2016.
\newblock \href {http://arxiv.org/abs/1608.06902} {{Recurrent Neural Networks
  With Limited Numerical Precision}}.
\newblock \emph{arXiv e-prints}, page arXiv:1608.06902.

\bibitem[{{Ott} et~al.(2018){Ott}, {Edunov}, {Grangier}, and
  {Auli}}]{2018arXiv180600187O}
Myle {Ott}, Sergey {Edunov}, David {Grangier}, and Michael {Auli}. 2018.
\newblock \href {http://arxiv.org/abs/1806.00187} {{Scaling Neural Machine
  Translation}}.
\newblock \emph{arXiv e-prints}, page arXiv:1806.00187.

\bibitem[{Park et~al.(2018)Park, Boo, Choi, Shin, and Sung}]{NIPS2018_8261}
Jinhwan Park, Yoonho Boo, Iksoo Choi, Sungho Shin, and Wonyong Sung. 2018.
\newblock Fully neural network based speech recognition on mobile and embedded
  devices.
\newblock In S.~Bengio, H.~Wallach, H.~Larochelle, K.~Grauman, N.~Cesa-Bianchi,
  and R.~Garnett, editors, \emph{Advances in Neural Information Processing
  Systems 31}, pages 10620--10630. Curran Associates, Inc.

\bibitem[{{Polino} et~al.(2018){Polino}, {Pascanu}, and
  {Alistarh}}]{2018arXiv180205668P}
Antonio {Polino}, Razvan {Pascanu}, and Dan {Alistarh}. 2018.
\newblock \href {http://arxiv.org/abs/1802.05668} {{Model compression via
  distillation and quantization}}.
\newblock \emph{arXiv e-prints}, page arXiv:1802.05668.

\bibitem[{{Rastegari} et~al.(2016){Rastegari}, {Ordonez}, {Redmon}, and
  {Farhadi}}]{2016arXiv160305279R}
Mohammad {Rastegari}, Vicente {Ordonez}, Joseph {Redmon}, and Ali {Farhadi}.
  2016.
\newblock \href {http://arxiv.org/abs/1603.05279} {{XNOR-Net: ImageNet
  Classification Using Binary Convolutional Neural Networks}}.
\newblock \emph{arXiv e-prints}, page arXiv:1603.05279.

\bibitem[{{See} et~al.(2016){See}, {Luong}, and
  {Manning}}]{2016arXiv160609274S}
Abigail {See}, Minh-Thang {Luong}, and Christopher~D. {Manning}. 2016.
\newblock \href {http://arxiv.org/abs/1606.09274} {{Compression of Neural
  Machine Translation Models via Pruning}}.
\newblock \emph{arXiv e-prints}, page arXiv:1606.09274.

\bibitem[{Sutskever et~al.(2014)Sutskever, Vinyals, and Le}]{NIPS2014_5346}
Ilya Sutskever, Oriol Vinyals, and Quoc~V Le. 2014.
\newblock Sequence to sequence learning with neural networks.
\newblock In Z.~Ghahramani, M.~Welling, C.~Cortes, N.~D. Lawrence, and K.~Q.
  Weinberger, editors, \emph{Advances in Neural Information Processing Systems
  27}, pages 3104--3112. Curran Associates, Inc.

\bibitem[{{Tang} and {Kwan}(1993)}]{229903}
C.~Z. {Tang} and H.~K. {Kwan}. 1993.
\newblock \href {https://doi.org/10.1109/78.229903} {Multilayer feedforward
  neural networks with single powers-of-two weights}.
\newblock \emph{IEEE Transactions on Signal Processing}, 41(8):2724--2727.

\bibitem[{{Tierno}(2019)}]{15848474}
Andrew {Tierno}. 2019.
\newblock \href
  {https://web.stanford.edu/class/cs224n/reports/custom/15848474.pdf}
  {{Quantized Transformer}}.
\newblock Technical report, Stanford University, Stanford, California.

\bibitem[{{Vaswani} et~al.(2017){Vaswani}, {Shazeer}, {Parmar}, {Uszkoreit},
  {Jones}, {Gomez}, {Kaiser}, and {Polosukhin}}]{transformer}
Ashish {Vaswani}, Noam {Shazeer}, Niki {Parmar}, Jakob {Uszkoreit}, Llion
  {Jones}, Aidan~N. {Gomez}, Lukasz {Kaiser}, and Illia {Polosukhin}. 2017.
\newblock \href {http://arxiv.org/abs/1706.03762} {{Attention Is All You
  Need}}.
\newblock \emph{arXiv e-prints}, page arXiv:1706.03762.

\bibitem[{Wang et~al.(2018)Wang, Xie, Deng, Li, Wang, and Xie}]{NIPS2018_7341}
Peiqi Wang, Xinfeng Xie, Lei Deng, Guoqi Li, Dongsheng Wang, and Yuan Xie.
  2018.
\newblock Hitnet: Hybrid ternary recurrent neural network.
\newblock In S.~Bengio, H.~Wallach, H.~Larochelle, K.~Grauman, N.~Cesa-Bianchi,
  and R.~Garnett, editors, \emph{Advances in Neural Information Processing
  Systems 31}, pages 604--614. Curran Associates, Inc.

\bibitem[{{Wen} et~al.(2017){Wen}, {He}, {Rajbhandari}, {Zhang}, {Wang}, {Liu},
  {Hu}, {Chen}, and {Li}}]{2017arXiv170905027W}
Wei {Wen}, Yuxiong {He}, Samyam {Rajbhandari}, Minjia {Zhang}, Wenhan {Wang},
  Fang {Liu}, Bin {Hu}, Yiran {Chen}, and Hai {Li}. 2017.
\newblock \href {http://arxiv.org/abs/1709.05027} {{Learning Intrinsic Sparse
  Structures within Long Short-Term Memory}}.
\newblock \emph{arXiv e-prints}, page arXiv:1709.05027.

\bibitem[{{Wu} et~al.(2015){Wu}, {Leng}, {Wang}, {Hu}, and
  {Cheng}}]{2015arXiv151206473W}
Jiaxiang {Wu}, Cong {Leng}, Yuhang {Wang}, Qinghao {Hu}, and Jian {Cheng}.
  2015.
\newblock \href {http://arxiv.org/abs/1512.06473} {{Quantized Convolutional
  Neural Networks for Mobile Devices}}.
\newblock \emph{arXiv e-prints}, page arXiv:1512.06473.

\bibitem[{{Wu} et~al.(2016){Wu}, {Schuster}, {Chen}, {Le}, {Norouzi},
  {Macherey}, {Krikun}, {Cao}, {Gao}, {Macherey}, {Klingner}, {Shah},
  {Johnson}, {Liu}, {Kaiser}, {Gouws}, {Kato}, {Kudo}, {Kazawa}, {Stevens},
  {Kurian}, {Patil}, {Wang}, {Young}, {Smith}, {Riesa}, {Rudnick}, {Vinyals},
  {Corrado}, {Hughes}, and {Dean}}]{2016arXiv160908144W}
Yonghui {Wu}, Mike {Schuster}, Zhifeng {Chen}, Quoc~V. {Le}, Mohammad
  {Norouzi}, Wolfgang {Macherey}, Maxim {Krikun}, Yuan {Cao}, Qin {Gao}, Klaus
  {Macherey}, Jeff {Klingner}, Apurva {Shah}, Melvin {Johnson}, Xiaobing {Liu},
  {\L}ukasz {Kaiser}, Stephan {Gouws}, Yoshikiyo {Kato}, Taku {Kudo}, Hideto
  {Kazawa}, Keith {Stevens}, George {Kurian}, Nishant {Patil}, Wei {Wang},
  Cliff {Young}, Jason {Smith}, Jason {Riesa}, Alex {Rudnick}, Oriol {Vinyals},
  Greg {Corrado}, Macduff {Hughes}, and Jeffrey {Dean}. 2016.
\newblock \href {http://arxiv.org/abs/1609.08144} {{Google's Neural Machine
  Translation System: Bridging the Gap between Human and Machine Translation}}.
\newblock \emph{arXiv e-prints}, page arXiv:1609.08144.

\bibitem[{{Zafrir} et~al.(2019){Zafrir}, {Boudoukh}, {Izsak}, and
  {Wasserblat}}]{2019arXiv191006188Z}
Ofir {Zafrir}, Guy {Boudoukh}, Peter {Izsak}, and Moshe {Wasserblat}. 2019.
\newblock \href {http://arxiv.org/abs/1910.06188} {{Q8BERT: Quantized 8Bit
  BERT}}.
\newblock \emph{arXiv e-prints}, page arXiv:1910.06188.

\bibitem[{{Zhang} et~al.(2018){Zhang}, {Yang}, {Ye}, and
  {Hua}}]{2018arXiv180710029Z}
Dongqing {Zhang}, Jiaolong {Yang}, Dongqiangzi {Ye}, and Gang {Hua}. 2018.
\newblock \href {http://arxiv.org/abs/1807.10029} {{LQ-Nets: Learned
  Quantization for Highly Accurate and Compact Deep Neural Networks}}.
\newblock \emph{arXiv e-prints}, page arXiv:1807.10029.

\bibitem[{{Zhou} et~al.(2016){Zhou}, {Wu}, {Ni}, {Zhou}, {Wen}, and
  {Zou}}]{2016arXiv160606160Z}
Shuchang {Zhou}, Yuxin {Wu}, Zekun {Ni}, Xinyu {Zhou}, He~{Wen}, and Yuheng
  {Zou}. 2016.
\newblock \href {http://arxiv.org/abs/1606.06160} {{DoReFa-Net: Training Low
  Bitwidth Convolutional Neural Networks with Low Bitwidth Gradients}}.
\newblock \emph{arXiv e-prints}, page arXiv:1606.06160.

\bibitem[{{Zhu} and {Gupta}(2017)}]{2017arXiv171001878Z}
Michael {Zhu} and Suyog {Gupta}. 2017.
\newblock \href {http://arxiv.org/abs/1710.01878} {{To prune, or not to prune:
  exploring the efficacy of pruning for model compression}}.
\newblock \emph{arXiv e-prints}, page arXiv:1710.01878.

\end{thebibliography}
\bibliographystyle{acl_natbib}

\clearpage

\appendix

\section{FullyQT Visual Guide} \label{sec:visual_guide}
\begin{figure*}
\centering
\begin{subfigure}{.65\textwidth}
    \centering
    \includegraphics[width=0.76923076923\linewidth]{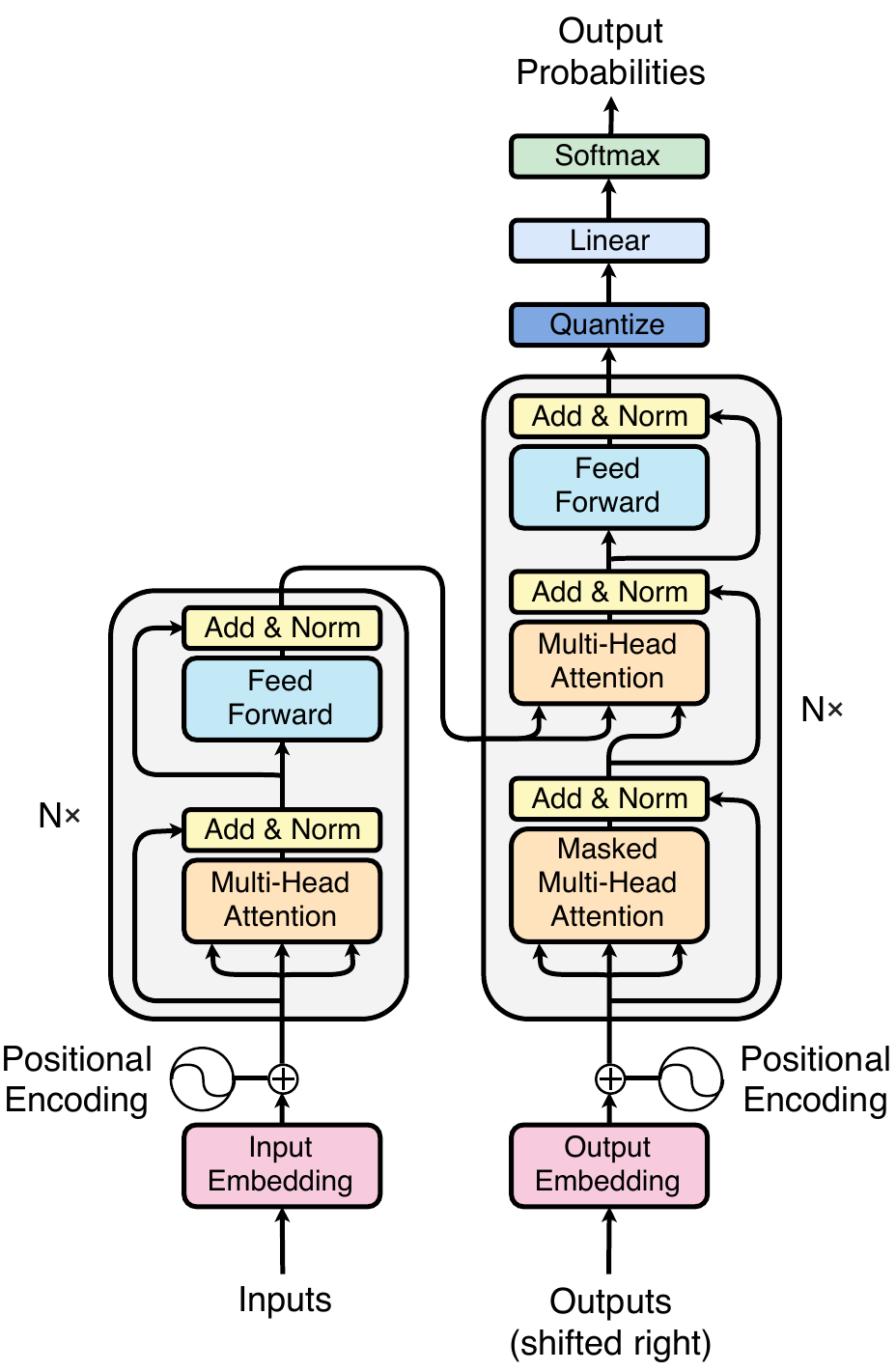}
\end{subfigure}%
\begin{subfigure}{.35\textwidth}
    \centering
    \includegraphics[width=0.3\linewidth]{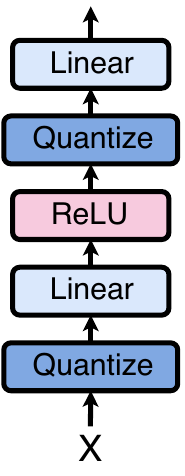}
\end{subfigure}
\caption{\label{fig:fullyqt_and_feedforward}
(left) Fully Quantized Transformer, (right) Feed-forward.}
\end{figure*}

\begin{figure*}
\centering
\begin{subfigure}{.5\textwidth}
    \centering
    \includegraphics[width=0.7\linewidth]{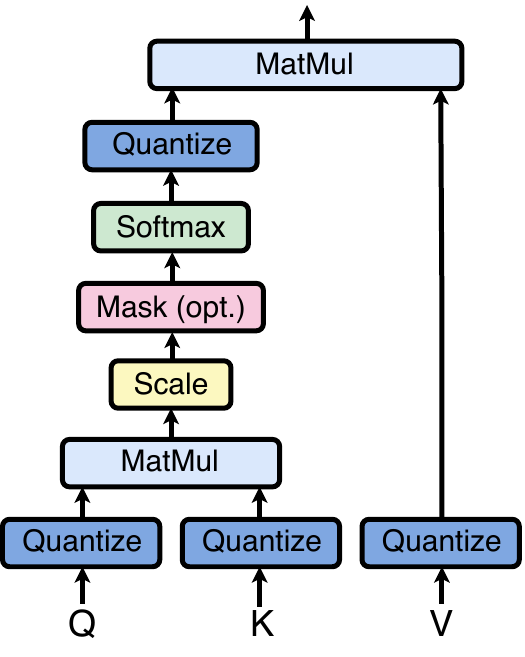}
\end{subfigure}%
\begin{subfigure}{.5\textwidth}
    \centering
    \includegraphics[width=0.7\linewidth]{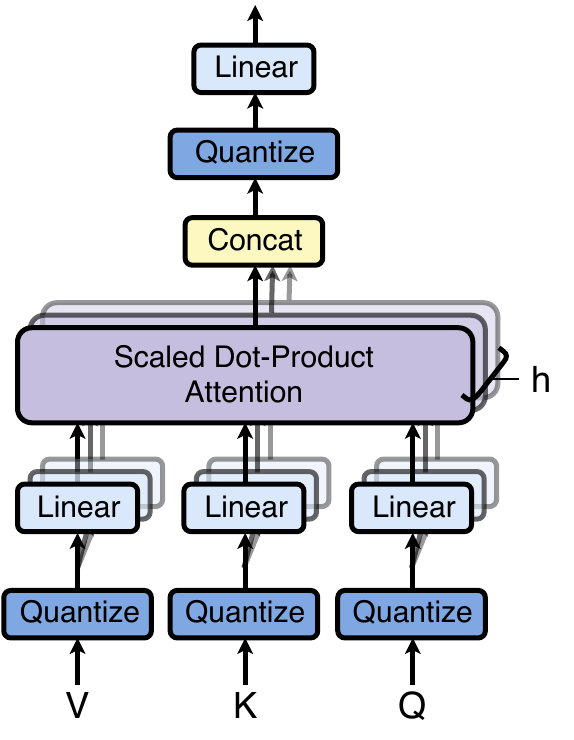}
\end{subfigure}
\caption{\label{fig:attention}
(left) Scaled Dot-Product Attention, (right) Multi-Head Attention.}
\end{figure*}

See Figure \ref{fig:fullyqt_and_feedforward} and \ref{fig:attention}.

\begin{table*}[!ht]
  \centering
  \resizebox{0.7\textwidth}{!}{
  \begin{tabular}{ccccccccc}
  \toprule
  \multirow{2}{*}{Model} & \multirow{2}{*}{Method} & \multirow{2}{*}{Precision} & \multicolumn{2}{c}{EN-CS} & \multicolumn{2}{c}{RU-EN} & \multicolumn{2}{c}{ES-EN} \\
  & & & PPL & BLEU & PPL & BLEU & PPL & BLEU \\
  \midrule
  \multirow{2}{*}{Base} & Baseline & 32-bit & 6.90 & 22.71 & 3.56 & 32.62 & 5.59 & \textbf{29.99} \\
  & FullyQT & 8-bit & 6.81 & \textbf{23.06} & 3.53 & \textbf{33.08} & 5.60 & 29.88 \\
  \midrule
  \multirow{2}{*}{Big} & Baseline & 32-bit & 7.41 & 22.22 & 3.57 & \textbf{32.22} & 5.32 & 30.06 \\
  & FullyQT & 8-bit & 7.17 & \textbf{22.49} & 3.66 & 31.74 & 5.35 & \textbf{30.15} \\
  \bottomrule
  \end{tabular}
  }
  \caption{\label{tab:additional-datasets}
  Evaluation of our quantization method on the WMT14 EN-CS, WMT14 RU-EN and WMT14 ES-EN translation datasets.}
\end{table*}

\section{Additional Datasets} \label{sec:additional_datasets}
We evaluated our quantization method on additional translation datasets (see Table \ref{tab:additional-datasets}). All models are trained following the same setup as in section \ref{sec:full_quantization}. Vocabulary size is set to 32k for all models. Since there is no test set for WMT14 ES-EN, we used the validation set as a test set and omitted computing any validation epochs during training.

\section{Pruning Useless Nodes}
We propose an additional compression method for the Transformer, which is independent of our quantization method. Both can be used conjointly to further compress the Transformer.

Once the model is fully trained and quantized, we can further compress it by removing useless nodes. By useless, we mean nodes which do not cause any loss in translation quality when removed. We choose to prune nodes instead of independently pruning weights. The latter method usually requires special hardware or software to leverage sparse weight matrices. Pruning nodes results in concretely shrunken models. When getting rid of a node, we remove its corresponding set of weights from the layer outputting it and the following layer receiving the node as input.

The only nodes of the Transformer which can be removed without causing alterations to other components of the network are the nodes in between the two layers of each feed-forward network. Fortunately, these consist of a substantial portion of the model's weights. In the case of the base Transformer, for a respective vocabulary of size 37000 and 32000, 39.96\% and 41.65\% of the total weights are owned by the feed-foward networks. This number grows to 47.03\% and 48.18\% in the big Transformer.

\begin{figure*}
\centering
\includegraphics[width=\linewidth]{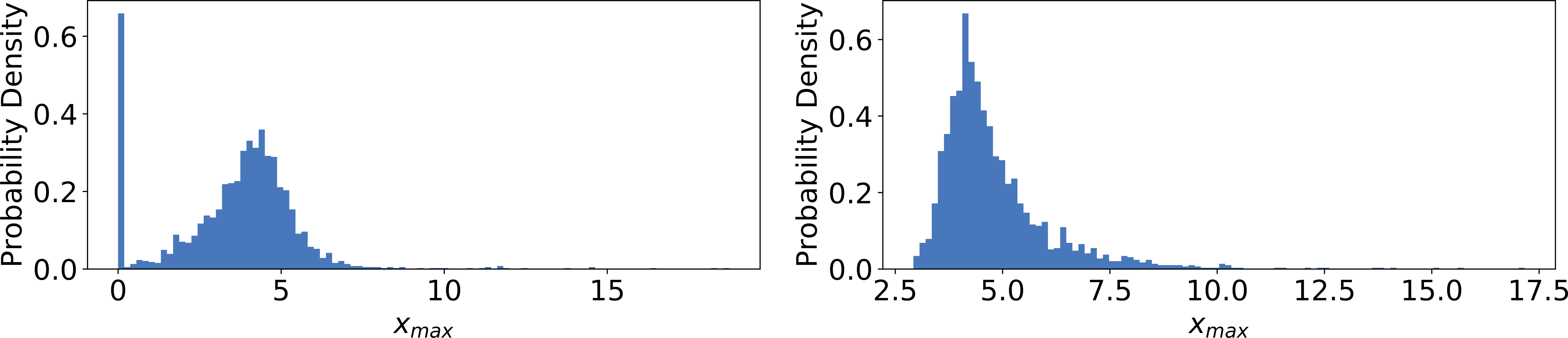}
\caption{\label{running_max_hist}
$x_{max}$ histogram of a ReLU layer in the encoder (left) and decoder (right), one $x_{max}$ per output node.}
\end{figure*}

To evaluate which nodes can be safely pruned without affecting translation quality, we estimate $x_{max}$ for each node of the ReLU output over a few hundred steps. This is done on the training set, using the fully trained model and keeping all other weights frozen. These $x_{max}$ are computed before quantizing the ReLU output and do not replace the ones used by the quantization process. Figure \ref{running_max_hist} shows the histogram of these running estimates for one ReLU layer in the encoder and one in the decoder. All other ReLU layers share the same pattern, where in the encoder there are always multiple $x_{max}$ close to 0. This does not happen in the decoder.

Once the running estimates are computed, we prune its corresponding node if $x_{max} < z \sigma$ where $z$ is a hyperparameter and $\sigma$ the standard deviation of the layer's $x_{max}$. We empirically found $z=0.025$ to work well, with higher thresholds causing BLEU to quickly decay. No retraining of the model is performed after pruning nodes.

Using this method, we can further compress the Transformer without affecting BLEU scores. Our approach has the advantage of being adaptive, meaning the number of nodes pruned per layer will differ as opposed to a fixed pruning ratio method. For example, in the case of the big Transformer trained on WMT14 EN-FR, 169 nodes were pruned in the first ReLU of the encoder, while in the second, 1226 were pruned. Nodes in the decoder rarely got pruned, at most 4 in the whole decoder. Results are presented in Table \ref{pruning}.

Other approaches usually decide the ratio first and then prune. We compared with two such methods. For each task, we fix their ratio to the average percentage of nodes pruned by our method and only prune nodes in the encoder. The first fixed pruning method uses L1-norm to sort nodes in ascending weight order, while the second sorts the $x_{max}$, also in ascending order.

Since $x_{max}$ is a running estimate, results varied per trial. However, the method only takes a few hundred training steps to perform. Running many trials is not an issue. Reported results are averaged on the test set over a few trials. BLEU varied by about 0.01$-$0.02.

\begin{table*}
  \centering
  \resizebox{\textwidth}{!}{
\begin{tabular}{ccc|cccc|cccc}
  \toprule
  \multirow{3}{*}{Model} & \multirow{3}{*}{Precision} & \multirow{3}{*}{Method} & \multicolumn{4}{c}{EN-DE} & \multicolumn{4}{c}{EN-FR} \\
  & & & \multirow{2}{*}{PPL} & \multirow{2}{*}{BLEU} & Nodes Pruned & Total & \multirow{2}{*}{PPL} & \multirow{2}{*}{BLEU} & Nodes Pruned & Total \\
  & & & & & in Encoder FF & Compr. & & & in Encoder FF & Compr. \\
  \midrule
  Base & 8-bit & No pruning & 4.39 & 27.60 & 0\% & 3.95x & 2.90 & 39.91 & 0\% & 3.95x \\
  & & L1-norm fixed & 5.57 & 23.99 & 13.57\% & 4.02x & 4.38 & 29.01 & 9.47\% & 3.99x \\
  & & $x_{max}$ fixed & 4.57 & 27.33 & 13.57\% & 4.02x & 3.18 & 39.40 & 9.47\% & 3.99x \\
  & & $x_{max}$ adaptive & 4.40 & \textbf{27.60} & 13.57\% & 4.02x & 2.90 & \textbf{39.91} & 9.47\% & 3.99x \\
  \cmidrule{2-11}
  & 6-bit & No pruning & 5.09 & 26.98 & 0\% & 5.25x & 3.38 & 37.07 & 0\% & 5.24x \\
  & & L1-norm fixed & 6.97 & 20.81 & 12.06\% & 5.31x & 4.19 & 31.64 & 9.62\% & 5.28x \\
  & & $x_{max}$ fixed & 5.41 & 26.20 & 12.06\% & 5.31x & 3.68 & 36.91 & 9.62\% & 5.28x \\
  & & $x_{max}$ adaptive & 5.09 & \textbf{26.98} & 12.06\% & 5.31x & 3.38 & \textbf{37.07} & 9.62\% & 5.28x \\
  \midrule
  Big & 8-bit & No pruning & 4.24 & 27.95 & 0\% & 3.97x & 2.80 & 40.17 & 0\% & 3.97x \\
  & & L1-norm fixed & 5.80 & 22.65 & 26.39\% & 4.21x & 4.16 & 28.85 & 28.41\% & 4.24x \\
  & & $x_{max}$ fixed & 4.47 & 27.43 & 26.39\% & 4.21x & 2.91 & 39.40 & 28.41\% & 4.24x \\
  & & $x_{max}$ adaptive & 4.25 & \textbf{27.95} & 26.39\% & 4.21x & 2.80 & \textbf{40.17} & 28.41\% & 4.24x \\
  \cmidrule{2-11}
  & 6-bit & No pruning & 4.78 & 26.76 & 0\% & 5.28x & 2.87 & 39.59 & 0\% & 5.28x \\
  & & L1-norm fixed & 7.73 & 17.32 & 29.96\% & 5.64x & 7.88 & 15.09 & 22.66\% & 5.54x \\
  & & $x_{max}$ fixed & 4.92 & \textbf{26.86} & 29.96\% & 5.64x & 2.91 & 39.25 & 22.66\% & 5.54x \\
  & & $x_{max}$ adaptive & 4.78 & 26.76 & 29.96\% & 5.64x & 2.87 & \textbf{39.59} & 22.66\% & 5.54x \\
  \bottomrule
  \end{tabular}
  }
  \caption{\label{pruning}
  Comparison of our adaptive pruning scheme versus fixed rate pruning methods for equal pruning proportions. Total compression accounts for quantization combined with pruning.}
\end{table*}

\end{document}